\documentclass[10pt,twocolumn,letterpaper]{article}

\usepackage{cvpr}
\usepackage{times}
\usepackage{epsfig}
\usepackage{graphicx}
\usepackage{amsmath}
\usepackage{amssymb}
\usepackage{gensymb}
\usepackage{siunitx}
\usepackage{textcomp}
\usepackage{multirow}
\usepackage{indentfirst}
\usepackage{authblk}
\usepackage{threeparttable}
\graphicspath{ {images/} }
\usepackage{graphicx}
\newcolumntype{P}[1]{>{\centering\arraybackslash}p{#1}}

\usepackage[pagebackref=true,breaklinks=true,letterpaper=true,colorlinks,bookmarks=false]{hyperref}

\cvprfinalcopy 

\def\cvprPaperID{3} 

\ifcvprfinal\fi
\begin{document}

\title{Scene Understanding Networks for Autonomous Driving based on Around View Monitoring System}

\renewcommand\footnotemark{}
\author[1,*]{JeongYeol Baek\thanks{*\:These two authors contributed equally to this work.}}
\author[2,*]{Ioana Veronica Chelu}
\author[2]{Livia Iordache}
\author[2]{Vlad Paunescu}
\author[1]{HyunJoo Ryu}
\author[2]{Alexandru Ghiuta}
\author[2]{Andrei Petreanu}
\author[1]{YunSung Soh}
\author[2]{Andrei Leica}
\author[1,\dag]{ByeongMoon Jeon\thanks{\dag\: E-mail: bm.jeon@lge.com}}

\affil[1]{Convergence Center, LG Electronics, Korea}
\affil[2]{Arnia Software, Romania}


\maketitle
\thispagestyle{empty}

\begin{abstract}
   Modern driver assistance systems rely on a wide range of sensors (RADAR, LIDAR, ultrasound and cameras) for scene understanding and prediction. These sensors are typically used for detecting traffic participants and scene elements required for navigation. In this paper we argue that relying on camera based systems, specifically Around View Monitoring (AVM) system has great potential to achieve these goals in both parking and driving modes with decreased costs. The contributions of this paper are as follows: we present a new end-to-end solution for delimiting the safe drivable area for each frame by means of identifying the closest obstacle in each direction from the driving vehicle, we use this approach to calculate the distance to the nearest obstacles and we incorporate it into a unified end-to-end architecture capable of joint object detection, curb detection and safe drivable area detection. Furthermore, we describe the family of networks for both a high accuracy solution and a low complexity solution. We also introduce further augmentation of the base architecture with 3D object detection.
\end{abstract}

\section{Introduction}

Visual environment perception plays a key role in the development of autonomous vehicles, providing fundamental information on the driving scene, including free space area and surrounding obstacles. These perception tasks can gather information from various sensors - LIDARs, cameras, RADARs or a fusion of them. Dense laser scanners are capable of creating a dynamic three-dimensional map of the environment and are best-suited for the task. However, their costs are still very high to be integrated in reasonably priced vehicles. Driven by the latest advances in the field of computer vision, we propose using only camera-based systems, which have the potential to reach dense laser-scan performance with lower cost. In particular, deep learning has fueled an improvement in accuracy of classic object detection and segmentation at an accelerated rate. However, object detection systems alone are usually not sufficient for autonomous emergency-braking and forward-collision systems, since the variety of possible road obstacles (e.g. tree branches, small animals) and road structure (e.g. country roads, different textures) make it impractical to train only typical object detection networks and road segmentation networks for scene understanding.  To tackle these problems, we present two main contributions:

\begin{itemize}
\item A new solution for delimiting the closest obstacles in all directions of the driving vehicle through bottom point estimation and curb detection, while also determining the exact distance to the nearest obstacles in each direction.
\item Integrating the obstacle detection network into a unified end-to-end solution capable of jointly delimiting the free drivable area by means of obstacle bottom point estimation, curb detection  and 2D multi-scale object detection for a low complexity solution.
\end{itemize}

\begin{figure}[t]
\begin{center}
\includegraphics[width=0.9\linewidth]{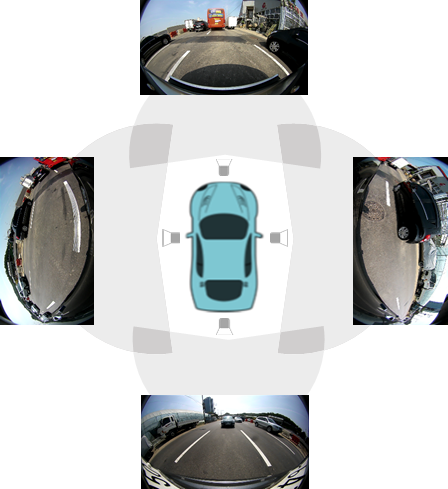}
\end{center}
   \caption{AVM camera system}
\label{fig:deepthinq}
\end{figure}

Scene understanding systems require high accuracy to ensure safety. However, model deployment on embedded platforms calls for real-time inference speed for prompt control and small model size for power-efficiency. We address these requirements by developing a low complexity solution which uses a light encoder network, benefits from sharing computations (i.e sharing the encoder) amongst the proposed perception tasks and uses single shot detection. We demonstrate the viability of our unified solution by showing that it achieves 16.7 fps on the Nvidia TX2 embedded platform.

Since 2D object detection sometimes provides insufficient information for scene understanding, we need to augment current solutions with 3D information to find the exact pose of objects in the 3D world. We propose augmenting the high accuracy solution to detect the orientation and dimension of each object.

This paper mainly focuses on detecting fundamental scene information for safe driving: object detection, curb detection, free drivable area segmentation, object distance from the camera and object orientation. We test our approach on a novel dataset consisting of fisheye images. A fisheye image is a wide-angle and distorted image which is generally used for \textit{Around View Monitoring} (AVM) systems (Figure~\ref{fig:deepthinq}). 
The advantage of using fisheye cameras in the field of autonomous driving is obvious, as they offer a much wider field of view (\ang{190}) than conventional cameras, thus capturing more relevant information of the car's surroundings (pedestrians, obstacles, etc.).
The four cameras are positioned at the front, rear, left, and right side of the vehicle and give drivers a \ang{360} view of their surroundings so as to check for obstructions around the vehicle.

\section{Related Work}

In this section we present a review on recent approaches for the tasks that we explore in the rest of the paper, i.e. object detection, classification, free space segmentation and 3D orientation.

\textbf{Free space detection: } State-of-the-art methods for detecting drivable area surface usually frame the problem in terms of road segmentation. Fully Convolutional Networks (FCN) \cite{fcn} use a convolutional network to perform spatially dense prediction tasks like semantic segmentation using transposed convolutions to model upsample layers. Later, dilated convolutions \cite{dilated} were also introduced to augment the receptive field of the network. The existing research generally tackles pixel segmentation networks or depth map derivation using stereo cameras. With these methods, unclassified pixels require complex post-processing to handle them. In this paper we propose using a simpler architecture for detecting obstacles and free space detection by identifying the bottom points of each obstacle in all directions of the driving vehicle.

\textbf{Object detection: } Modern neural network approaches to object detection can be divided into two categories: region proposal based methods and single-shot methods. The former category covers approaches like Faster R-CNN \cite{faster} that have a two-step process which involves first generating region proposals using an RPN (region proposal network) and then scoring them using a secondary module. In the single-shot network approach \cite{ssd}, the region proposal and classification stages are integrated into one single stage, by using predefined anchor boxes (priors) like a sliding window that moves through each spatial position on the feature map to concurrently predict bounding boxes and class confidence scores. Performing region proposal and classification network simultaneously makes this approach extremely fast in comparison with two-stage methods.

\textbf{3D Object detection: } 3D object detection has gathered significant consideration lately due to its key contribution in applications that require interactions with objects in real-world scenarios, as in autonomous driving.  This issue has been addressed from a purely geometric point of view (\eg estimating the pose of an object with 6DoF from a single image), as well as using DCNNs (deep convolutional neural networks) in order to reconstruct 3D models. In \cite{Rothganger}, Rothganger \etal  use local affine-invariant image descriptors in order to construct 3D models of object instances in 2D images and then matching them with 3D poses in the image. In \cite{Hara2017DesigningDC}, Hara \etal. demonstrate DCNN effectiveness in estimating the \ang{0} to \ang{360} orientation of objects. Mousavian \etal \cite{Mousavian} use a DCNN to regress stable 3D object features, while other methods exploit depth information from stereo images \cite{3DOP-stereo}, or combine temporal information using structure from motion algorithms in order to augment 2D detections with 3D information.


\section{Networks for Scene Understanding}
In this section we give a detailed description of network architectures which we propose for AVM scene understanding, including object detection, free/drivable area segmentation, object distance and object orientation. For 2D object detection we investigate standard object detection networks such as Faster R-CNN and SSD \cite{ssd}. We also experiment with different encoders on our AVM dataset, such as MobileNet \cite{mobilenet} and Inception-ResNet-V2 \cite{inc-res-v2}, plugging them in the standard object detection network to evaluate the accuracy and the runtime performance. Free space detection is achieved by finding the bottom point of each pixel column in the image. The union of all the bottom points in an image represents the bottom boundary of all obstacles in the scene and all the pixels beneath it are considered free space. The distance to the nearest obstacle is obtained mathematically by applying camera geometry to the bottom points of an object. In addition, a multi-task network architecture is proposed to jointly perform object detection and bottom point localization for the low complexity solution. Detecting object orientation of each vehicle  is achieved using a 3D object detection network. Integration of 3D object detection into the low complexity solution is left as future work.

\subsection{Bottom-Net}

\begin{figure}[t]
\begin{center}
\includegraphics[width=0.9\linewidth]{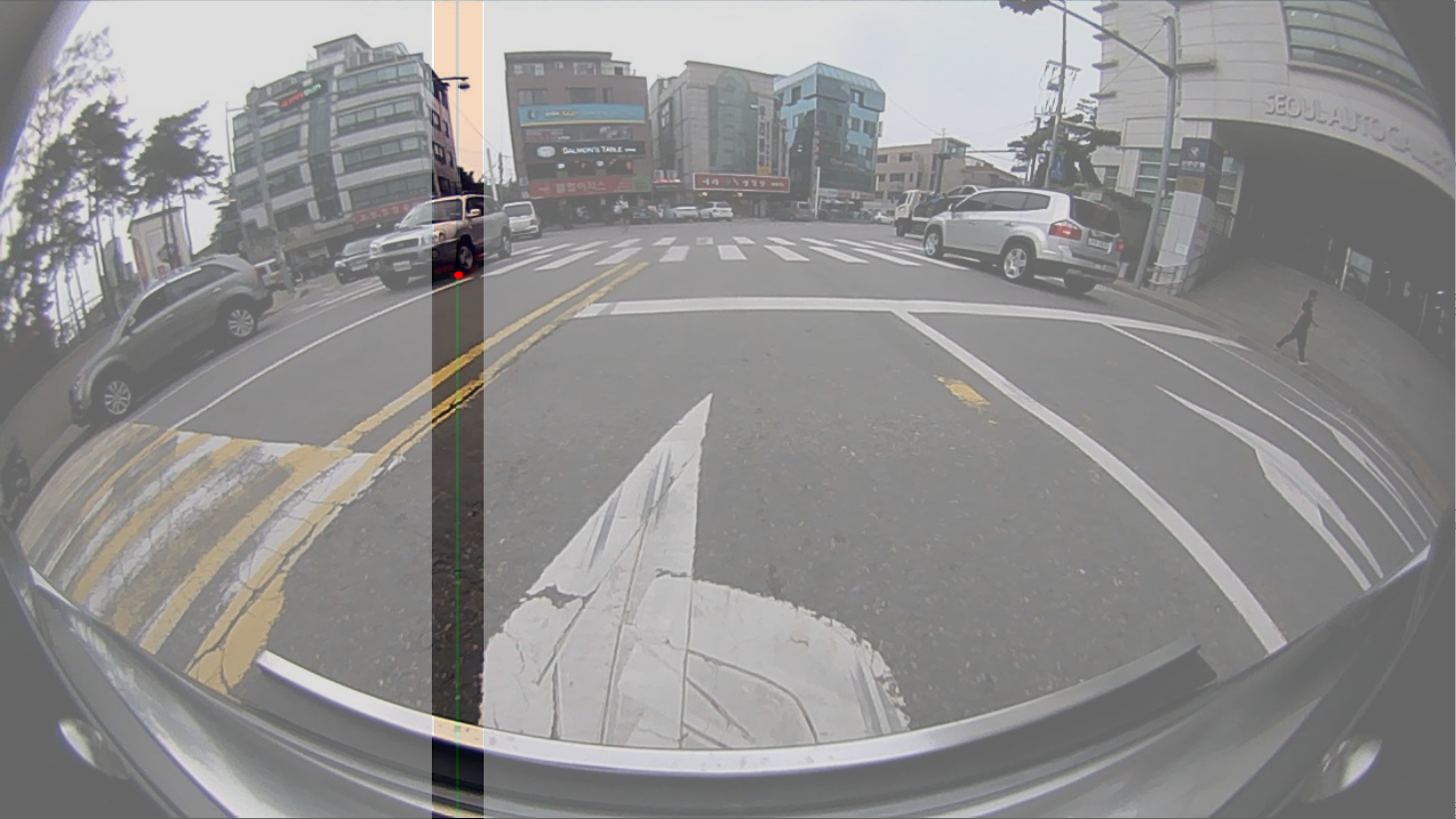}
\end{center}
   \caption{A columnwise prediction for a corresponding pixel augmented with the adjacent 24 pixels area}
\label{fig:stixel}
\end{figure}

Our approach for detecting the curb and the free drivable area is inspired by a stixel representation of the world \cite{stixel-world} \cite{stixelnet}. Originally, the network takes as input each vertical column of an image. The input columns that the network used had width $24$, overlapped over $23$ pixels like in Figure~\ref{fig:stixel}. Each column would then be passed through a convolutional network to output one-of-k labels, with k being the height dimension. As a result, it would learn to classify the position of the bottom pixel of the obstacle corresponding to that column. The union of all columns would build either the curb or the free drivable area of the scene.
\begin{figure}[t]
\begin{center}
\includegraphics[width=0.9\linewidth]{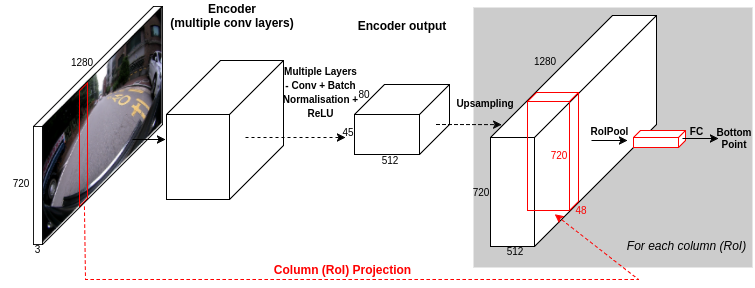}
\end{center}
   \caption{Bottom prediction architecture using ROI pooling for each column}
\label{fig:q-bottom-roi}
\end{figure}

In this architecture, due to the overlapping between the columns, more than 95$\%$ of the computation is redundant. Motivated by this observation we replaced the columnwise network implementation with an end-to-end architecture that would accept a whole image as input. This network encoded the image into a deep feature map using multiple convolutional layers and then used multiple upsampling layers to generate a feature map having the same resolution as the input image. Inspired by the region of interest (ROI) approach of object detection systems, we cropped hardcoded regions of the image corresponding to the pixel columns augmented with the neighboring area of 23 pixels. As a result, the regions of interest for cropping the upsampled feature map are 23 pixels wide and 720 (height) pixels tall. We slide this window horizontally over the image at each x-coordinate. The resulting crops are then resized to a fixed length (e.g. 7x7) in the ROI pooling layer and are then classified to one-of-k classes (k is 720 in the high-accuracy case, i.e. the height of the image), in order to ultimately predict the bottom point. An illustration of the architecture is presented in Figure~\ref{fig:q-bottom-roi}. 
\begin{figure*}
\begin{center}
\includegraphics[width=.9\linewidth]{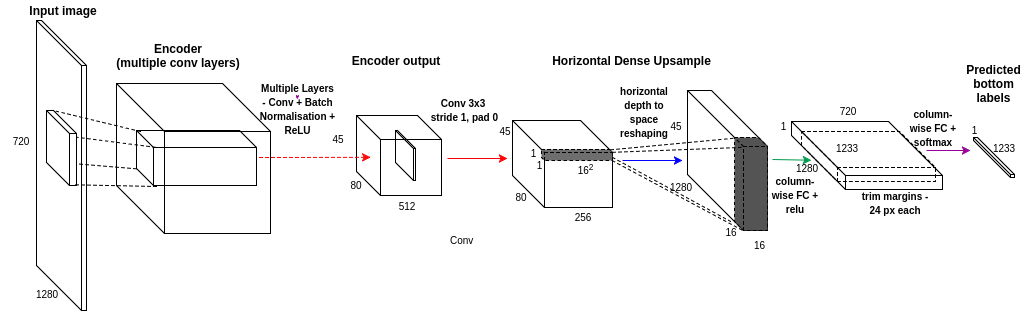}
\end{center}
   \caption{Bottom-Net architecture}
\label{fig:q_bottom_slim}
\end{figure*}

However, using ROIs with fixed positions for the final classification leads to repeated computation due to the overlap in the regions of interest. This insight naturally brings us to the final version of the bottom prediction network, which is to use a single shot method for the final classification layer of the bottom prediction task. Moreover, to make the network more efficient, we also replace the decoder part of the network corresponding to the multiple upsample layers with a single dense horizontal upsampling layer \cite{duc}. The resulting feature map generated from the encoder after applying multiple convolutions with $stride > 1$ has a resolution of $[width/16, height/16]$, being reduced 16 times the original image size. Compared with the previous version of the bottom prediction network, which used standard upsampling layers in both horizontal and vertical directions, the final upsampling method now generates output feature maps of size $[width, height/16]$ having their width multiplied $16\times$, leaving the height unchanged. The performance comparison between the standard and the proposed upsampling method along with details of each are reported in Section 5.2.1.

Finally, we add another fully connected layer on top of the horizontal upsampling layer to make a linear combination of each column's input. A softmax is used to classify each of the resulted columns to one-of-k categories, where k is the height of the image being predicted (in the high-accuracy case 720). Each column classification subtask automatically takes into account the pixels displayed in the proximity of the center column being classified and represents the final bottom prediction. Figure~\ref{fig:q_bottom_slim} depicts the final architecture of the bottom pixel prediction task.

\subsection{OD-Net}

The initial architecture we investigated is based on Faster R-CNN and predicts the bounding box and class of the objects in the scene. This architecture, combined with deep and powerful encoders like Inception-ResNet-V2, tends to offer the most accurate models as the high accuracy solution, but falls short of real-time performance on embedded systems.

To ensure the responsiveness of the low complexity solution for embedded systems, we need an effective object detection system which directly outputs object class probabilities and bounding box coordinates. We combine single shot detection with a light encoder like MobileNet for fast inference.

\subsection{Unified-Net} 

Unified architectures which combine the bottom prediction and the object detection networks usually take advantage of shared computation of the network encoder for better training optimization and runtime performance. Thus, we consider branching off the shared encoder at different layers so as to find the best trade-off between runtime performance and accuracy, details of which can be found in Section 5.2.4.

\begin{figure}[t]
\begin{center}
\includegraphics[width=0.9\linewidth]{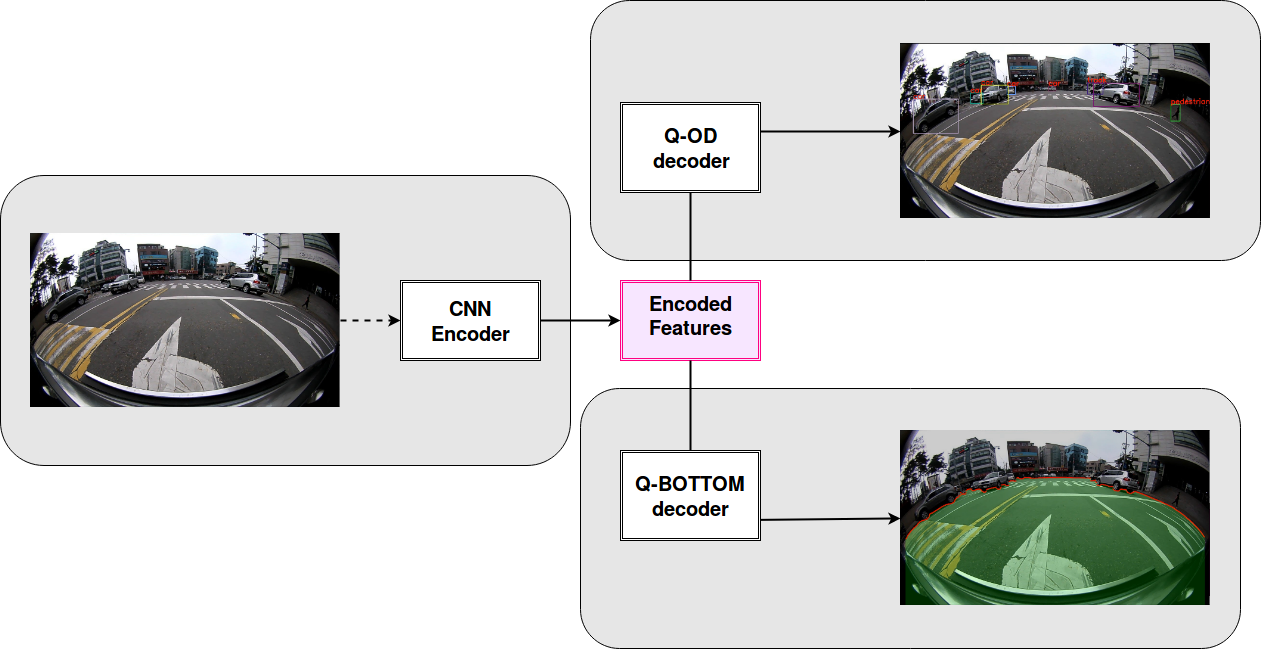}
\end{center}
   \caption{Unified-Net architecture}
\label{fig:unified}
\end{figure}

The unified architecture we propose encodes an image using a convolutional network and uses multiple decoders for each of the tasks. The unified architecture is illustrated in Figure~\ref{fig:unified}.

\textcolor{black}{
Branching at the input layer means that the two networks do not share any computation and we report results for this architecture as an upper bound in terms of accuracy. Branching off higher in the network lets us share computation between the two tasks and achieve better inference time, but lower accuracy, since the two heads have to share the feature representation in the encoder. 
The high level features of the convolutional encoder are slightly different for bottom prediction and object detection. As a result, branching off at a lower layer in the encoder increases accuracy and lets the two heads specialize their features. }
\subsection{3D-Net} 

\begin{figure}[t]
\begin{center}
\includegraphics[width=0.9\linewidth]{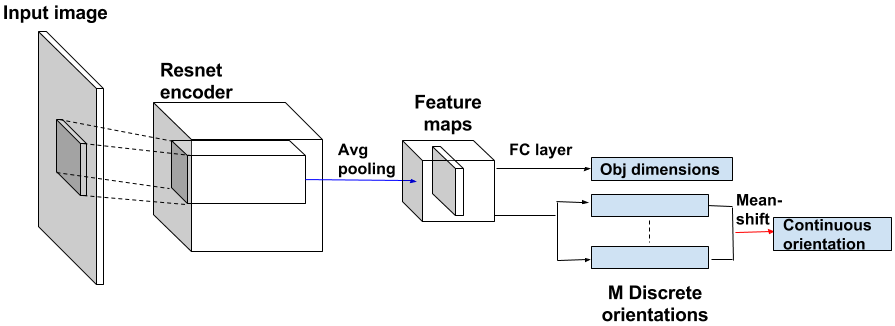}
\end{center}
   \caption{3D-Net architecture}
\label{fig:3dod-arch}
\end{figure}

In this section we present the high accuracy solution of orientation and dimension problem. We use ResNet-101 \cite{DBLP:journals/corr/HeZRS15} (the top 22 residual blocks) for the underlying DCNN architecture, as depicted in \cite{Hara2017DesigningDC}, pretrained on a subset of ImageNet with 1000 classes \cite{ILSVRC15}. The final architecture consists of two branches, for object orientation estimation based on angle discretization and for object dimensions regression, respectively. The network architecture is illustrated in Figure~\ref{fig:3dod-arch}.

The 3D network takes as input crops of the objects and estimates the real-world dimensions and orientation for each one of them. The 2D crops are extracted using OD-net and reprojected from the fisheye projection to a Lambert Cylindrical Equal-Area projection\cite{lamb-eq}. The Lambert projection function can be described as:

\begin{equation}
\begin{aligned}
L_{p}(\vec{R}) = \left[ \begin{array}{c} \lambda_{l}(\vec{R}) \\ \sin(\phi_{l}(\vec{R})) \end{array}\right],
\end{aligned}
\end{equation}
where $ \lambda_{l} $ and $ \phi_{l} $ are the latitude and longitude of a given ray $ \vec{R} $: 
\begin{equation}
\begin{aligned}
\lambda_{l}(\vec{R}) = \arccos(\frac{\vec{R_{y}} \cdot \left[\begin{array}{ccc}0 & 0 & 1\end{array} \right]^{T}}{|\vec{R_{y}}|})
\end{aligned}
\end{equation}
\begin{equation}
\begin{aligned}
\phi_{l}(\vec{R}) = \arccos(\frac{\vec{R_{y}} \cdot \vec{R}}{|\vec{R_{y}}| \cdot |\vec{R}|})
\end{aligned}
\end{equation}
Here, $ \vec{R_{y}} $ is the three dimensional projection of $ \vec{R} $ onto $ XoZ $:
\begin{equation}
\begin{aligned}
\vec{R_{y}} = \left[ \begin{array}{ccc} 1 & 0 & 0 \\ 0 & 0 & 0 \\ 0 & 0 & 1 \end{array} \right] \cdot \vec{R}
\end{aligned}
\end{equation}
Given the fisheye projection $ \vec{p} $ of a ray, we can compute $ \vec{R} $ as follows: 
\begin{equation}
\begin{aligned}
\vec{R} = \left[ \begin{array}{c} p_{x} \\ p_{y} \\ \frac{|\vec{p}|}{\tan(f_{p}^{-1}(|\vec{p}|))} \end{array} \right]
\end{aligned}
\end{equation}
In order to adjust for camera pitch, we rotated along $ oX $ with $ -\alpha $:
\begin{equation}
\begin{aligned}
\vec{R^{'}} = \left[ \begin{array}{ccc} 1 & 0 & 0 \\ 0 & \cos(-\alpha) & -\sin(-\alpha) \\ 0 & \sin(-\alpha) & \cos(-\alpha) \end{array} \right] \vec{R}
\end{aligned}
\end{equation}
Finally, the reprojected vector $ \vec{q} $ was computed as:
\begin{equation}
\begin{aligned}
\vec{q} = L_{p}(\vec{R^{'}})
 \end{aligned}
\end{equation}

\textcolor{black}{
The network predicts the object dimensions and object orientation. We transform the orientation prediction into a classification problem by discretizing the orientation value into N unique orientations. We subsequently recover the continuous orientation by using the mean-shift algorithm.}

\section{Implementation details}

In this section we describe the various training details we employ in our unified architectures.

\textbf{Preprocessing: } As dataset augmentation, we use random color distortion (brightness, saturation, contrast, hue) and normalization between $[0, 1]$. We also extend the training set by using horizontal flip. The low complexity solution also performs random cropping.

\textbf{Objective functions: } Bottom-Net uses a softmax cross-entropy loss for classifying each bottom point of each obstacle in all directions of the driving vehicle : 
\begin{equation}
\begin{aligned}
L_{bottom}(p, q) = -\dfrac{1}{w} \dfrac{1}{h}\sum_{k=0}^{w} \sum_{c=0}^{h} q_k(c)log p_k(c),
\end{aligned}
\end{equation}
where \textit{w} and \textit{h} are the width and the height of each frame.
Unified-Net sums up a total of three objective functions: one classification loss for the bottom pixel prediction and two losses for the detection task, classification and bounding box regression as detailed in \cite{faster}.

\textbf{Metrics: }
For the object detection task we use the classic mean average precision (mAP) metric at 0.5 intersection over union (IoU) overlap. For the bottom pixel prediction task we introduce the mean absolute error (MAE), which represents the mean pixel displacement between the ground truth and bottom pixel prediction.
We reuse the MAE metric for the 3D-OD task to compute object orientation and for object dimensions on all 3 axes.

\textbf{Low complexity solution details: }
Runtime performance and memory footprint are critical for real-time applications like the ones required for autonomous driving or for driver assistance systems. The embedded solution we propose for this requirement uses the MobileNet encoder and solves the two related tasks of object detection and bottom pixel detection. 

For the real-time embedded system used for prompt vehicle control, computational efficiency is more important, so the Unified-Net branches off at a higher convolutional layer of the encoder. Regarding the object detection task, we choose to use a multi-scale single shot network due to its fast runtime performance. For the encoder we use a trimmed version of MobileNet. We have found that eliminating the last 2 convolutional layers in the MobileNet encoder gave better accuracy. \textcolor{black}{
We detect objects at 6 scales, using the last encoder layer ("conv11") as the first feature map and create the remaining 5 as an extension to the encoder. Each feature map is responsible for detecting objects at different scales.
}



The training procedure for the Unified-Net uses 640x360 pixel resolution images trained in batches of 8. We use an initial learning rate of $7e-4$ and decay it every $10000$ iterations from the total number of $40000$ iterations.

\textbf{High accuracy solution details: } For the best possible accuracy we use the top part of the Inception-ResNet-V2 architecture as the encoder, with weights pretrained on ImageNet, for both bottom prediction and object detection. For the object detection task, we choose the Faster R-CNN architecture, which provides the best localization and classification accuracy.

Training is performed at full size resolution: 1280x720 with a batch size of 1, whilst keeping the same training procedure as in the original implementation. 

\section{Experiments}

In this section we provide details on the fisheye AVM dataset that we use for training and report the experimental evaluation we performed on it.

\textbf{AVM dataset } In order to evaluate the accuracy and runtime performance of our solution we constructed a dataset comprised of images taken using the AVM camera. This technology captures fisheye images, i.e. images with \ang{190} field of view, from 4 cameras, placed on different sides of the car. The fisheye camera captures more information from the car's surroundings. The dataset consists of driving footage captured on streets, parking lots, and highways. The car used is an SUV equipped with 4 fisheye cameras placed on the front grill, the rear bumper, and two side mirrors. For training and validation only the front camera images were used. The images are Full High Definition (FHD) and are split into train and validation set with a ratio of 8:2. Unfortunately, the dataset we used is not publicly available at this point.

The dataset consists of 2213 images containing 5 classes: car, bus, truck, pedestrian and *-cycle (bicycle/motorcycle). We use 1744 images for training and the rest 469 for testing. The classes are highly imbalanced, with the car class having a ratio of approximately 6:1 to each of the other classes as detailed in Table~\ref{db_stats} .
\begin{table}
\begin{center}
\begin{tabular}{l|l|l|l|l|l}
\textbf{Split} & \textbf{car} & \textbf{bus}  & \textbf{truck} & \textbf{pedestrian} & \textbf{*-cycle}\\\hline
Train & 5948  &669 &858 &1130 &320  \\
Test &1132 &271 &414 &121 &50        
\end{tabular}
\end{center}
\caption{Dataset statistics: number of objects in class}
\label{db_stats}
\end{table}

For the bottom pixel prediction task we use 3994 images for training and 601 for testing with the ground truth information outlined as y coordinates between $[0, h]$ (where \textit{h} is the height of each image) for each of the x coordinates between $[24, 1256]$.

\begin{figure}[t]
\begin{center}
\includegraphics[width=0.9\linewidth]{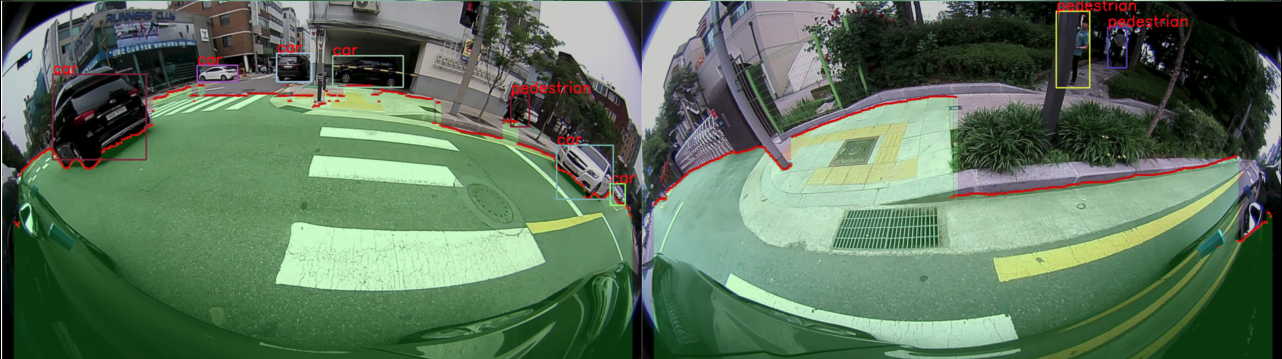}
\end{center}
   \caption{Side view detections. (left) left view. (right) right view.}
\label{fig:sides}
\end{figure}

Despite using only front-view images for training and testing, applying the detection system to the side view images produces good results as is depicted in Figure ~\ref{fig:sides}.

We tested the network accuracy and runtime under both scenarios – with original AVM images and with reprojected images. The accuracy improvement is negligible when we use the reprojected images, but the runtime performance increases due to the preprocessing step. Empirically, we noticed the convolutional network encoder performs well even in spite of the fact that the input image is distorted. 

\textbf{Embedded systems targeted platforms }
The embedded platform is Nvidia Jetson TX2. Tests for runtime performance on this platform use JetPack 3.1, Tensorflow 1.3, CUDA 8.0 with cuDNN 6.0.

\subsection{Performance comparison}

In this section we summarize the experimental results of the high accuracy solution and the low complexity solution.

\begin{table*}
\begin{center}
\resizebox{\linewidth}{!}{
\begin{threeparttable}
\begin{tabular}{p{1.7cm}|l|P{3.2cm}|l|l|l|l|l|l|l|l}
\textbf{Architecture}         & \textbf{Input size} & \textbf{Encoder}          & \textbf{Car}  & \textbf{Bus}  & \textbf{Truck} & \textbf{Pedestrian} & \textbf{*-cycle} & \textbf{mAP}  & \textbf{MAE}   & \textbf{TX2 fps} \\\hline
Unified-Net LC \tnote{1} & 640x360    &  MobileNet 0.5 & 0.68 & 0.75 & 0.38  & 0.30       & 0.58    & \textbf{0.55} & \textbf{2.77} \tnote{3}  &  \textbf{16.7} \\
Unified-Net HA \tnote{2} & 1280x720   &  Inception-ResNet-V2 & 0.91 & 0.84 & 0.65  & 0.62       & 0.85    & \textbf{0.77} & \textbf{3.7} \tnote{3} & -  \\
OD-Net  & 1280x720   &  Inception-ResNet-V2 & 0.87 & 0.88 & 0.70  & 0.72       & 0.81    & \textbf{0.80} & - & -   \\
Bottom-Net &  1280x720   &  Inception-ResNet-V2 & - & - & -  & -       & -    & - & \textbf{3.7} & -  \\
\end{tabular}
\begin{tablenotes}
      \small
      \item[1] low complexity solution
      \item[2] high accuracy solution
      \item[3] MAE of full resolution input \vs resized input is not directly comparable, since the former adds mean errors from 2$\times$ the input pixels of the latter.
\end{tablenotes}
\end{threeparttable}}
\end{center}
\caption{Performance comparison of Unified-Net LC, Unified-Net HC, OD-Net and Bottom-Net}
\label{accuracy_results}
\end{table*}

\textbf{High accuracy solution: } We report individual results for OD-Net and Bottom-Net in terms of both accuracy and runtime performance. Performance results for the Unified-Net in the high accuracy solution are also depicted in Table~\ref{accuracy_results} in comparison to the low complexity solution results. In this experiment, we branched off the Inception-ResNet-V2 encoder after the \textit{PreAuxLogits} layer, corresponding to the first stage in the Faster R-CNN architecture.

\textbf{Low complexity solution: } 
For embedded systems, our investigation into which architecture is best suited for our Unified-Net yielded the MobileNet encoder with depth multiplier 0.5 and SSD decoder for the object detection task. Section 5.2.2 provides additional details into the encoder size ablation study experiments. The Unified-Net was tuned to achieve the best balance between mAP and MAE. We tested our unified architecture on the TX2 platform and report runtime performance and accuracy results in Table~\ref{accuracy_results}.

\begin{figure}[t]
\begin{center}
\includegraphics[width=0.9\linewidth]{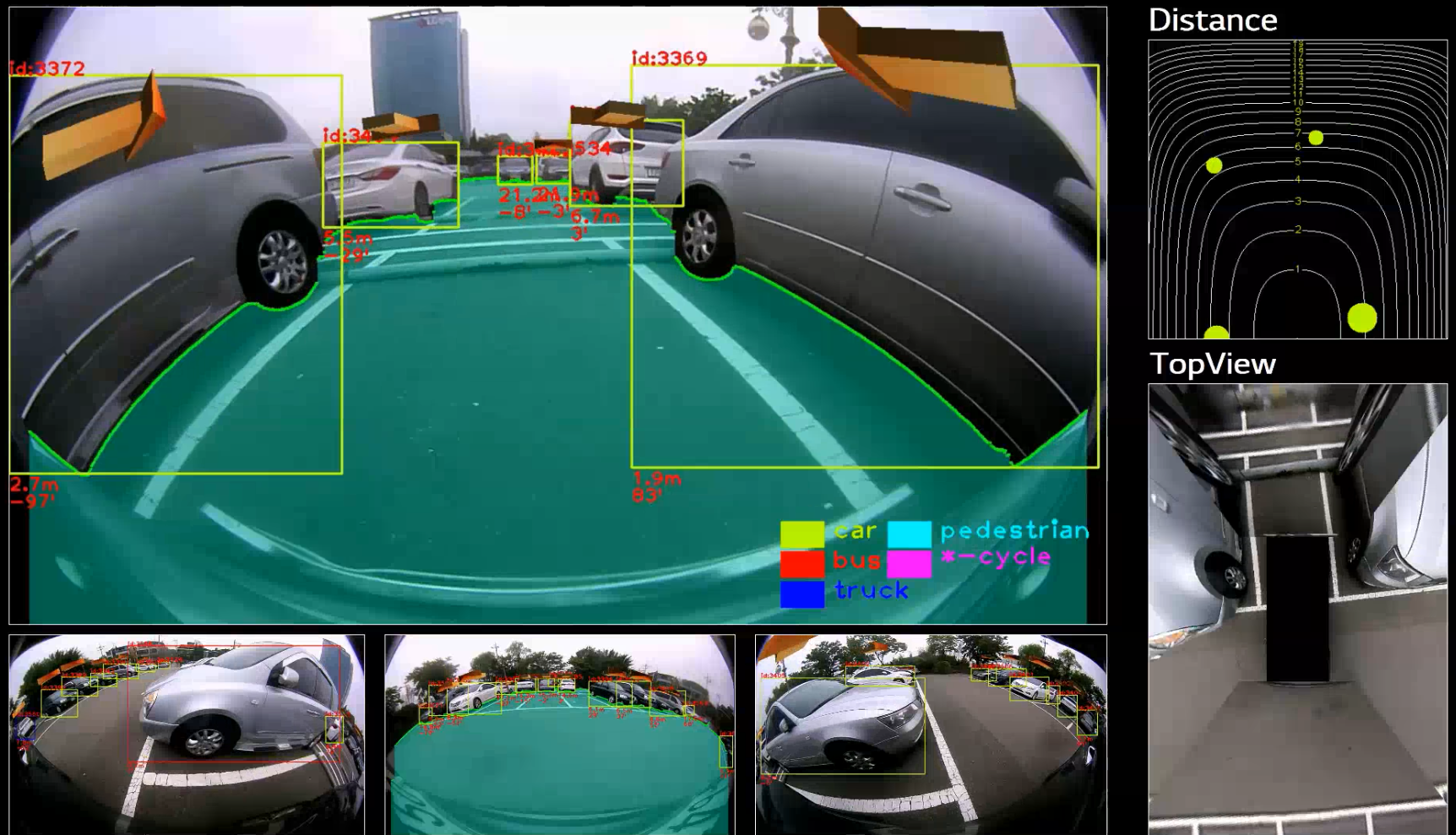}
\end{center}
   \caption{Captured frame from the high accuracy solution}
\label{fig:joint_solution}
\end{figure}

\textbf{3D high accuracy solution: } We report the orientation and dimension performances in Table~\ref{3dod_avm} and show the visual results in Figure~\ref{fig:joint_solution} in which the arrows represent object orientation.


\begin{table}
\begin{center}
\begin{threeparttable}
\begin{tabular}{c|P{2cm}|P{0.65cm}|P{0.65cm}|P{0.65cm}|P{0.65cm}|P{0.65cm}}
\multicolumn{1}{P{1cm}}{\textbf{Class}} & \multicolumn{1}{|P{1cm}}{\textbf{Orientation MAE ($^{\circ}$)}} & \multicolumn{3}{|c}{\textbf{Dims MAE (m)}} \\
\hline
& &  X &  Y &  Z \\ \hline
*cycle & 4.09 & 0.082 & 0.081 & 0.132  \\
car & 5.3 & 0.094 & 0.087 & 0.191  \\
truck & 4.84 & 0.112 & 0.126 & 0.251  \\
barrier & 2.01 & 0.164 & 0.039 & 0.098  \\
bus & 3.06 & 0.136 & 0.174 & 0.392 
\end{tabular}
\end{threeparttable}
\end{center}
\caption{Accuracy performance of 3D-Net}
\label{3dod_avm}
\end{table}

\subsection{Ablation studies}

We performed several investigations to increase the accuracy and reduce the runtime. In this section we detail them and report their performance in comparison with the baseline solution.

\subsubsection{Bottom-Net in high accuracy solution}

\begin{table}
\begin{center}
\begin{threeparttable}
\begin{tabular}{p{1.5cm}|P{3.4cm}|l|P{1.5cm}}
\textbf{Decoder} & \textbf{Encoder} & \textbf{MAE} & \textbf{TitanX fps} \\ \hline
FCN-A \tnote{1} & Inception-ResNet-V2 & 4.1 & 3.9 \\
FCN-B \tnote{2} & Inception-ResNet-V2 & 4.0 & 2.9 \\
HDUC \tnote{3}  & Inception-ResNet-V2  & 3.8 & 5.7 \\
HDUC  & Inception-ResNet-V2 \tnote{4} & \textbf{3.7} & \textbf{17.2} \\ 
\end{tabular}
\begin{tablenotes}
      \small
      \item[1] FCN-A: original FCN8s
      \item[2] FCN-B: skip layers 1,2,3,4 + upsample 2x + dilation trick
      \item[3] Horizontal Dense Upsample Convolution
      \item[4] Inception-ResNet-V2 up to PreAuxLogits layer
    \end{tablenotes}
    \end{threeparttable}
\end{center}
\caption{Accuracy and runtime performance of Bottom-Net}
\label{q_bot}
\end{table}

We report results on the comparison in terms of accuracy and runtime performance of the two main architectures we used for the decoder part. The networks use the Inception-ResNet-V2 encoder. Before using the dense upsampling layer, we tried using two versions of the FCN \cite{fcn} for the decoder, the original FCN8s and one version using skip layers from pool 1, 2, 3, 4, one upsample layer 2x and dilation trick for the last layer to increase the resolution of the feature map. The comparison in terms of runtime performance is performed on Nvidia Titan X and shown in Table ~\ref{q_bot}.

\subsubsection{Image resolution and encoder size in low complexity solution}
\begin{figure}[t]
\begin{center}
\includegraphics[width=0.9\linewidth]{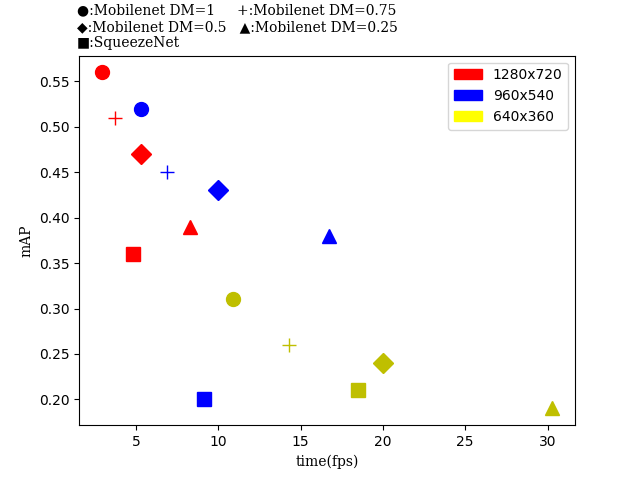}
\end{center}
   \caption{Accuracy \vs runtime performance for 1-scale architectures with different encoder sizes}
\label{fig:1_scale}
\end{figure}

We performed various experiments using different encoders and different input resolutions in order to find a balance between accuracy and inference time. The results in Figure~\ref{fig:1_scale}  are performed using SSD 1-scale (using only one feature map) because the training time is faster and allows for various experiments to be performed. The figure illustrates that full resolution and 960x540 resolution have similar results, while resolution 640x360 achieves lower accuracy, but faster inference time. For each resolution, the accuracy achieved with the encoder decreases as the depth multiplier (DM) for the MobileNet encoder is decreased.

\subsubsection{Unified-Net in low complexity solution}
\begin{table}
\begin{center}
\begin{threeparttable}
\resizebox{\linewidth}{!}{
\begin{tabular}{l|P{2cm}|l|l|l|l|l}
\textbf{Architecture} & \textbf{Branch-off} & \textbf{FPS TX2} & \textbf{MAE} & \textbf{mAP} \\ \hline
Unified-Net & conv2 & 16.4 & 2.85 & 0.51\\ 
Unified-Net & conv5 & 16.7 & \textbf{2.77} & \textbf{0.55}\\
Unified-Net & conv11  & 17.0 & 3.25 & 0.53\\
Unified-Net & conv11 (+conv12, +conv13) & 16.9 & 3.56 & 0.51\\
\end{tabular}}

    \end{threeparttable}
\end{center}
\caption{Unified-Net results for various branching layers}
\label{branching_results}
\end{table}

We investigated the accuracy \vs complexity trade-off between multiple versions of our embedded system solution as we vary the branching layer of the multi-task network. We experimented with branching at the "conv2", "conv5" and "conv11" layers with the trimmed MobileNet encoder and branching at "conv11" using the full encoder (including "conv12", "conv13") and report the results in terms of accuracy and runtime performance in Table~\ref{branching_results}. We choose as baseline model the branching at "conv5" since it achieves the best balance of MAE and fps.

\section{Conclusion}
In this paper we introduced a new way of detecting the free drivable area by means of bottom point estimation of obstacles in each direction of the driving vehicle and incorporated it into a multi-task unified architecture for a low complexity solution which enables curb detection, free drivable area segmentation, object detection and object distance, while achieving object localization and classification of obstacles pertaining to 5 classes: car, bus, truck, pedestrian and *-cycle. The proposed approach is capable of achieving 16.7 fps on the TX2 targeted embedded system.
We leave as future development the integration of 3D-Net into the low complexity solution. In the future we also aim to include instance segmentation into the unified system in order to improve the overall performance.

{\small
\bibliographystyle{ieee}
\bibliography{sample}
}

\end{document}